\documentclass{article}

\usepackage{arxiv}

\usepackage[utf8]{inputenc} 
\usepackage[T1]{fontenc}    
\usepackage{hyperref}       
\usepackage{url}            
\usepackage{booktabs}       
\usepackage{amsfonts}       
\usepackage{nicefrac}       
\usepackage{microtype}      
\usepackage{lipsum}
\usepackage{graphicx}
\graphicspath{ {./images/} }
\usepackage{algorithm}
\usepackage{algorithmic}

\title{UltrasODM: A Dual Stream Optical Flow Mamba Network for 3D Freehand Ultrasound Reconstruction}

\author{
  Mayank Anand\thanks{Corresponding author.\newline This paper has been accepted for presentation at AAAI 2026, Singapore, in the AIMedHealth workshop track.} \\
  Department of Information Technology \\
  Indian Institute of Information Technology Allahabad \\
  Prayagraj, India \\
  \texttt{anandmayank698@gmail.com} \\
  \And
  Ujair Alam \\
  Department of  Electronics \& Communication \\ Engineering\\
  Indian Institute of Information Technology Allahabad \\
  Prayagraj, India \\
  \And
  Surya Prakash \\
  Department of Electronics \& Communication \\ Engineering \\
  Indian Institute of Information Technology Allahabad \\
  Prayagraj, India \\
  \And
  Priya Shukla \\
  Department of Information Technology \\
  Indian Institute of Information Technology Allahabad \\
  Prayagraj, India \\
  \And
  Gora Chand Nandi \\
  Department of Information Technology \\
  Indian Institute of Information Technology Allahabad \\
  Prayagraj, India \\
  \And
  Dom\`enec Puig \\
  Department of Computer Engineering and Mathematics \\
  Universitat Rovira i Virgili \\
  Tarragona, Spain \\
}

\begin{document}
\maketitle
\begin{abstract}
Clinical ultrasound acquisition is highly operator-dependent, where rapid probe motion and brightness fluctuations often lead to reconstruction errors that reduce trust and clinical utility. We present \textit{UltrasODM}, a dual-stream framework that assists sonographers during acquisition through calibrated per-frame uncertainty, saliency-based diagnostics, and actionable prompts. UltrasODM integrates (i) a contrastive ranking module that groups frames by motion similarity, (ii) an optical-flow stream fused with Dual-Mamba temporal modules for robust 6-DoF pose estimation, and (iii) a Human-in-the-Loop (HITL) layer combining Bayesian uncertainty, clinician-calibrated thresholds, and saliency maps highlighting regions of low confidence. When uncertainty exceeds the threshold, the system issues unobtrusive alerts suggesting corrective actions such as re-scanning highlighted regions or slowing the sweep. Evaluated on a clinical freehand ultrasound dataset, UltrasODM reduces drift by $\mathbf{15.2\%}$, distance error by $\mathbf{12.1\%}$, and Hausdorff distance by $\mathbf{10.1\%}$ relative to UltrasOM, while producing per-frame uncertainty and saliency outputs. By emphasizing transparency and clinician feedback, UltrasODM improves reconstruction reliability and supports safer, more trustworthy clinical workflows. Our code is publicly available at \mbox{\url{https://github.com/AnandMayank/UltrasODM}}.
\end{abstract}


\section{Introduction}
UltraSound imaging (US) is preferred for diagnostics and interventions due to affordability, safety, portability, and real-time imaging. Compared to the 2D US, 3D US offers enhanced contextual information, is gaining traction in various applications, including anatomical measurements \cite{looney2018fully}, \cite{zhou2020voxel}, standard plane acquisition \cite{yang2021agent}, and interventional guidance \cite{li2015ultrasound}, \cite{zettinig2016toward}. According to the methodology used for data acquisition, 3D US systems can be categorized into 2D arrays, mechanical systems, and free-hand techniques. The 2D matrix and mechanical systems often suffer from high cost and limited field of view. Some of the free-hand imaging systems are also used in clinical applications such as endorectal procedures (mainly for prostate brachytherapy), cancer biopsy, and needle pathfinding \cite{bax2008mechanically}; \cite{wei2004robot}; \cite{yan2012automatic} including the use of optical or electromagnetic (EM) sensors to obtain spatial information from 2D images \cite{mozaffari2017freehand}. 3D freehand ultrasound reconstruction seeks to eliminate external tracking devices by deriving spatial relationships from image data alone. Although it is possible to affix an external positional device to the transducer, this results in an increase in the bulkiness of the transducer, resulting in the reduction of the image quality generated and frequently leading to imprecise measurements within a clinical setting due to various optical or electrical interferences \cite{luo2023recon}. This process entails estimating the 6 degrees-of-freedom (DoF) parameters to ascertain the 3D positions of 2D ultrasound images \cite{wang2022medical}. This methodology provides a more adaptable, convenient, and economical option for three-dimensional medical imaging \cite{yan2024fine} Nevertheless, accurately reconstructing 3D ultrasound video frames without pose information poses a considerable challenge.

This marks the advent of deep learning methodologies, wherein the autonomous reconstruction of 3D US imagery, devoid of reliance on extrinsic positioning apparatus, can be effectively employed. Recent studies were mainly based on convolutional neural network (CNN) and achieved advanced performance. Prevost et al. \cite{prevost20183d} introduced an end-to-end method utilizing CNN to estimate the relative motion of US images. Guo et al. \cite{guo2022ultrasound} proposed a deep contextual-contrastive network (DC2-Net) and applied a margin triplet loss for contrastive learning in a regression task to improve reconstruction
performance. Li et al. \cite{li2023long} proposed to estimate the 3D spatial transformation between US frames using recurrent neural networks (RNN). Luo et al. \cite{luo2021self}, \cite{luo2023multi} further improve reconstruction performance through online learning and shape priors. However, we note that the general approach of these studies is to first extract the coarse-grained features of the image and then extract the temporal information contained in these features. Recently, Luo et al. \cite{luo2023recon} introduced an online learning reconstruction framework (RecON), imposing constraints such as motion-weighted training loss, frame-level contextual consistency, and path-label similarity, which significantly improved the accuracy of motion estimation in complex scan motions. With the increasing popularity of transformers, Ning et al. \cite{lee2025enhancing} applied a transformer architecture to combine local and long-range information from a CNN-based backbone encoder and IMU sensors.

However, the challenge of accurately reconstructing 3D US images is exacerbated by rapid probe movements and significant brightness variations, which can overwhelm optical flow algorithms, leading to inaccuracies in motion estimation [1]. Models often struggle with back-and-forth scanning patterns, resulting in errors in reconstruction length calculations. Consequently, variations in ultrasonic scanning protocols, such as probe parameters and sampling frame rates, can adversely affect reconstruction quality, indicating potential generalization issues despite normalized preprocessing \cite{sun2025ultrasom}. 

In recent times, research has emerged advocating the utilization of Mamba, which has been shown to enhance the long-range dependency management capabilities of the state space model (SSM). Mamba \cite{gu2023mamba}, characterized as a proficient neural network architecture, has excelled in the fields of feature extraction and spatio-temporal information processing, achieving significant achievements in domains such as image classification, semantic segmentation, and video comprehension \cite{zhang2025mamba}, \cite{chen2024video}, \cite{chen2024rsmamba}. It adeptly captures global spatio-temporal correlations within sequences while preserving computational efficiency and resilience.

Motivated by Mamba's ability to extract textual features from sequences, this manuscript introduces a novel ultrasound reconstruction framework that synergistically integrates Optical Flow and the Mamba architecture, designated UltrasOM, to estimate of freedom (6DoF) parameters between ultrasound image sequences. It amalgamates the benefits of self-attention mechanisms for spatial feature extraction with Mamba's capacity to capture temporal information, while simultaneously incorporating dynamic features facilitated by optical flow. In response to these challenges, we propose UltrasODM: a dual Mamba architecture that replaces the traditional bidirectional Mamba approach. Our method leverages contrastive learning \cite{basu2022unsupervised} to sort frames based on marginal ranking, ensuring that frames with similar 6DoF parameters are processed together. This innovative strategy not only enhances the model's ability to handle complex scanning patterns but also improves the accuracy and stability of 3D ultrasound reconstructions. By integrating these advances, we aim to provide a more effective solution for clinical applications, ultimately enhancing the diagnostic capabilities of ultrasound imaging.

\section{Method}

In this section, we detail the architecture and workflow of the proposed \textit{UltrasODM} framework for 3D freehand ultrasound reconstruction. The methodology is designed to accurately estimate spatial transformations between ultrasound frames without external tracking. To this end, we present a dual-stream architecture that separately models global and local motion using farthest and nearest point sampling strategies. These sampled motion cues are processed through bidirectional Mamba networks for temporal modeling, integrated with optical flow for enhanced motion-aware feature extraction. The framework concludes with a multi-scale fusion strategy for 6-DoF parameter regression, supported by a multi-component clinical loss function. Each component of the proposed pipeline is described in detail in the following subsections.

\begin{figure*}[t]
  \centering
  \includegraphics[width=\textwidth, clip, trim=0 0 0 0]{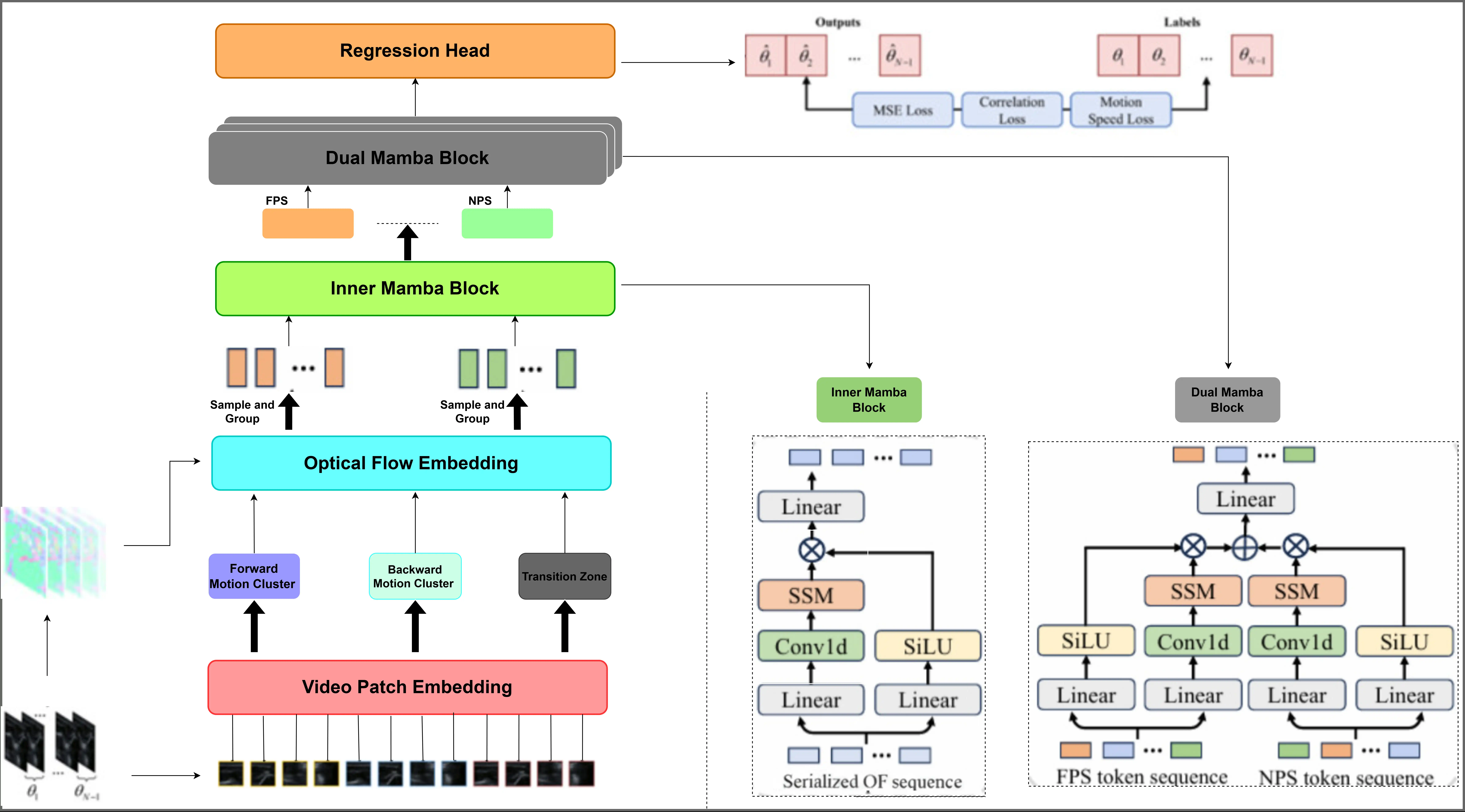}
  \caption{UltrasODM network architecture.}
  \label{fig:UltrasODM}
\end{figure*}

\subsection*{Problem Formulation and Dual-Stream Architecture}

An ultrasound sequence \(S\) comprises a set of temporally ordered frames \(S = \{I_m\},\ m = 1, 2, \dots, M\), acquired at increasing timestamps. For any frame pair \(I_i\) and \(I_j\), the relative spatial transformation is defined by a 6-DOF rigid motion vector \(t_{j \leftarrow i},\ 1 \leq i < j \leq M\), capturing the displacement between frames.

Conventional motion estimation methods typically treat all image regions uniformly, resulting in redundant computation and suboptimal feature utilization, particularly in the presence of heterogeneous anatomical structures and probe-induced motion. To address these limitations, \textit{UltrasODM} adopts a dual-stream design that separately models global and local motion cues using spatial sampling strategies. These motion-aware features are then encoded using bidirectional Mamba networks, enabling efficient temporal modeling of motion trajectories. The outputs are subsequently fused to regress accurate 6-DoF transformations between ultrasound frames.

\subsubsection*{Contrastive Frame Embedding for Motion Initialization}

Before applying optical flow estimation and dynamic grouping, we first embed each ultrasound frame into a motion-aware latent space using a contrastive learning objective. 
Rather than performing hard clustering or discarding frames at this stage, this process provides \textit{motion similarity priors} that guide downstream modules in identifying coherent motion phases.

Given a sequence of ultrasound frames \( S = \{ I_t \}_{t=1}^{T} \), each frame is mapped to an embedding vector 
\(\mathbf{e}_t = f_{\phi}(I_t)\) through a lightweight encoder \( f_{\phi} \) trained using a triplet loss:

\begin{equation}
\mathcal{L}_{\mathrm{tri}} =
\max \Bigl(
\| f_{\phi}(I_a) - f_{\phi}(I_p) \|_2^2
-\| f_{\phi}(I_a) - f_{\phi}(I_n) \|_2^2
+ \alpha,\; 0
\Bigr)
\label{eq:triplet}
\end{equation}

Here, temporally adjacent frames (\(|a-p|\le\delta\)) act as positives, while temporally distant frames (\(|a-n|\ge\Delta\)) serve as negatives. 
This encourages the network to learn embeddings where temporally consistent frames (i.e., similar motion directions) lie close together, while frames with distinct motion patterns are well-separated.

During inference, all frames are projected into this embedding space, forming natural clusters corresponding to \textit{forward}, \textit{backward}, and \textit{transition} motion phases. 
Instead of enforcing hard clustering early, these motion similarity embeddings are used as soft priors within the \textit{Dynamic Grouping} and \textit{Optical Flow Encoding} modules. 
This ensures that UltrasODM retains full temporal continuity while benefiting from contrastively learned motion cues that stabilize probe direction transitions and reduce ambiguity in motion estimation.

The following pseudocode (Algorithm~\ref{alg:contrastive_grouping}) outlines the procedure used for learning these embeddings. 
While the algorithm produces motion-coherent frame groupings using DBSCAN during visualization or analysis, in the full UltrasODM pipeline these groupings are not used to truncate the sequence. 
Instead, the similarity structure derived from the embeddings serves as an auxiliary signal for subsequent modules.

\begin{algorithm}[ht]
\caption{Contrastive Frame Grouping}
\label{alg:contrastive_grouping}
\begin{algorithmic}[1]
\STATE \textbf{Input:} Frame sequence $\{I_t\}_{t=1}^T$, margin $\alpha = 0.2$
\STATE \textbf{Initialize:} Embedding network $f_{\phi}$
\FOR{each training iteration}
    \STATE Sample anchor $I_a$, positive $I_p$ ($|a-p| \le \delta$), negative $I_n$ ($|a-n| \ge \Delta$)
    \STATE Compute embeddings: $\mathbf{e}_t = f_{\phi}(I_t)$
    \STATE Compute triplet loss: 
     \[
      \mathcal{L}_{\mathrm{tri}} = \max\bigl( \|\mathbf{e}_a - \mathbf{e}_p\|_2^2
      - \|\mathbf{e}_a - \mathbf{e}_n\|_2^2 + \alpha,\; 0 \bigr)
    \]
\ENDFOR
\STATE \textbf{Inference:}
\STATE Compute all embeddings: $\{\mathbf{e}_t\}_{t=1}^T$
\STATE Cluster embeddings using \texttt{DBSCAN}:
\STATE \quad $\{\mathcal{G}_k\}_{k=1}^K \leftarrow \mathrm{DBSCAN}(\{\mathbf{e}_t\},\,\epsilon=\tau_{\mathrm{sim}})$
\STATE \textbf{Output:} \raggedright Frame groups $\{\mathcal{G}_k\}$ representing motion-coherent frame clusters
\end{algorithmic}
\end{algorithm}

\subsection*{Dual-Stream Motion Sampling Strategy}

To capture both coarse and fine-grained motion information during ultrasound probe movement, the proposed \textit{UltrasODM} framework incorporates a dual-stream motion sampling strategy. This approach processes global and local motion cues in parallel using two complementary techniques: Farthest Point Sampling (FPS) for global motion and Nearest Point Sampling (NPS) for local motion. Each stream independently samples relevant features from the ultrasound frames, which are later fused to provide an accurate estimation of the 6-DoF transformation estimation.

\subsubsection*{Global Motion Stream via Farthest Point Sampling (FPS):}

The global motion stream employs Farthest Point Sampling (FPS) to extract spatially diverse points throughout the ultrasound image, enabling the model to capture large-scale probe displacements. Given a set of candidate points \(P\), FPS iteratively selects the next point that maximizes the minimum Euclidean distance from the previously selected subset \(S_i\), as defined below:
\begin{equation}
    p_{i+1} = \arg\max_{p \in P \setminus S_i} \min_{q \in S_i} \|p - q\|_2
\end{equation}

This strategy ensures wide spatial coverage while maintaining computational efficiency, making it suitable for modeling global motion patterns affecting the entire field of view.

\subsubsection*{Local Motion Stream via Nearest Point Sampling (NPS):}

\begin{table*}
\centering
\caption{Comparison of methods on quantitative metrics. The proposed Optical Flow + Mamba and UltrasODM demonstrate consistent improvements over existing baselines, with UltrasODM outperforming in most metrics.}
\label{tab:results}
\small
\setlength{\tabcolsep}{3pt} 
\resizebox{\textwidth}{!}{%
\begin{tabular}{lrrrrrrrr}
\hline
\textbf{Method} & \textbf{DE (mm)} & \textbf{FD (mm)} & \textbf{FDR (\%)} & \textbf{ADR (\%)} & \textbf{MD (mm)} & \textbf{SD (mm)} & \textbf{HD (mm)} & \textbf{MEA (deg)} \\
\hline
Linear               & 10.00(3.94) & 14.65(6.39) & 13.92(5.27) & 20.76(5.89) & 17.12(6.39) & 2695.51(988.68) & 14.82(6.26) & 2.50(1.89) \\
CNN-OF               & 8.71(3.77)  & 12.33(5.84) & 12.20(6.83) & 19.74(6.93) & 14.02(6.02) & 2376.34(941.74) & 12.45(5.83) & 2.30(1.74) \\
DC$^2$-net           & 8.64(3.32)  & 11.16(5.30) & 11.80(4.78) & 18.36(5.30) & 13.96(5.23) & 2098.20(806.04) & 11.21(5.30) & 2.19(1.75) \\
UltrasOM             & 7.34(3.11)  & 10.67(5.91) & \textbf{10.24(5.34)} & 17.15(4.60) & 13.15(5.50) & 1966.59(784.56) & 10.81(5.81) & 2.05(1.74) \\
\hline
Optical Flow + Mamba & 7.90(3.40)  & 10.95(5.60) & 10.60(5.10) & 17.05(4.55) & \textbf{13.10(5.30)} & 1940.00(770.00) & 10.60(5.70) & 2.02(1.70) \\
\textbf{UltrasODM (ours)}     & \textbf{7.12(3.08)} & \textbf{10.41(5.45)} & 10.35(5.28) & \textbf{16.85(4.42)} & 13.20(5.48) & \textbf{1895.20(748.50)} & \textbf{10.39(5.64)} & \textbf{1.99(1.68)} \\
\hline
\end{tabular}}
\end{table*}

To complement the global stream, the local motion stream focuses on anatomically informative regions using Nearest Point Sampling (NPS). High-gradient regions are first identified based on a threshold \(\tau_{\mathrm{grad}}\), which isolates areas with significant motion cues:
\begin{equation}
    R_{\mathrm{high}} = \{p \in P : \|\nabla I(p)\| > \tau_{\mathrm{grad}}\}
\end{equation}

From each detected region, the method samples \(k\) nearest neighbors around local maxima, concentrating attention on tissue interfaces, boundaries, and other salient anatomical features. This allows the model to prioritize clinically relevant motion patterns while reducing computational overhead in homogeneous areas.

\subsection*{Inner + Dual Mamba Temporal Modeling}
To effectively capture temporal dependencies within ultrasound sequences, \textit{UltrasODM} employs a two-stage Mamba architecture composed of an \textbf{Inner Mamba} for local causal modeling and a \textbf{Dual Mamba} for global context aggregation. 
This structure replaces the previous bidirectional Mamba design, allowing the network to jointly encode fine-grained temporal continuity and long-range spatial consistency.

\paragraph{Inner Mamba: Local Causal Encoding.}
Given the motion-aware frame embeddings $\{f_t\}_{t=1}^{T}$ obtained after feature extraction and optical flow fusion, 
the Inner Mamba models short-term temporal dynamics within a causal window $\mathcal{W}_t = \{f_{t-k}, \dots, f_t\}$ using a linear state-space update approach:
\begin{equation}
\mathbf{h}_t = \mathbf{A}\mathbf{h}_{t-1} + \mathbf{B}f_t, \qquad
\mathbf{y}_t^{(\mathrm{in})} = \mathbf{C}\mathbf{h}_t + \mathbf{D}f_t,
\end{equation}
where $\mathbf{A}, \mathbf{B}, \mathbf{C}, \mathbf{D}$ are learnable parameters.
This causal formulation preserves the inherent forward acquisition order of ultrasound data while capturing local motion variations and probe--tissue interactions.
The resulting embeddings $\{\mathbf{y}_t^{(\mathrm{in})}\}_{t=1}^{T}$ provide temporally coherent representations of local motion patterns.

\paragraph{Dual Mamba: Global Aggregation.}
To capture long-range dependencies and maintain structural continuity across the sequence, 
the local embeddings from the Inner Mamba are serialized under two complementary token orderings:
\begin{equation}
\mathbf{Z}_F = \{\mathbf{y}_{\pi_F(i)}^{(\mathrm{in})}\}_{i=1}^{M}, \qquad
\mathbf{Z}_N = \{\mathbf{y}_{\pi_N(i)}^{(\mathrm{in})}\}_{i=1}^{M},
\end{equation}
where $\pi_F$ corresponds to Farthest Point Sampling (FPS) emphasizing global coverage, and $\pi_N$ corresponds to Nearest Point Sampling (NPS), which emphasizes local continuity.
Each sequence is independently processed by a dedicated Mamba encoder:
\begin{equation}
    \mathbf{O}_F = \mathrm{Mamba}_{\mathrm{FPS}}(\mathbf{Z}_F; \Theta_F)
\end{equation}
and
\begin{equation}
\mathbf{O}_N = \mathrm{Mamba}_{\mathrm{NPS}}(\mathbf{Z}_N; \Theta_N)
\end{equation}

producing two complementary global representations.
A gating-based fusion mechanism integrates the outputs as
\begin{equation}
    \mathbf{O}_{\mathrm{dual}} =
\begin{array}[t]{l}
\sigma(\mathbf{W}_g[\mathbf{O}_F,\mathbf{O}_N]) \odot \mathbf{O}_F
+ (1 - \sigma(\mathbf{W}_g[\mathbf{O}_F,\mathbf{O}_N])) \odot \mathbf{O}_N,
\end{array}
\end{equation}

where $\sigma(\cdot)$ denotes a sigmoid activation and $\odot$ element-wise multiplication.
The fused representation $\mathbf{O}_{\mathrm{dual}}$ encodes both the global spatial layout and the temporal continuity of the ultrasound sequence. This $\mathbf{O}_{\mathrm{dual}}$ is utilized within the feature integration and 6-DoF regression module to infer spatial transformations by frame.

\subsection*{Feature Integration and 6-DOF Estimation}

To regress accurate 6-DOF spatial transformations between ultrasound frames, \textit{UltrasODM} integrates features from both global and local motion streams using a multi-scale fusion strategy. Features are extracted via an adapted EfficientNet-B1 backbone \cite{tan2019efficientnet}, enhanced with video patch embedding and 3D positional encoding to preserve spatiotemporal relationships:
\begin{equation}
    e_{t,i,j} = \mathrm{Linear}(\mathrm{Patch}(f_{t,i,j})) + \mathrm{PE}(t, i, j)
\end{equation}

In parallel, optical flow priors are estimated and incorporated to guide motion-aware representation learning. These priors are fused with the learned features using an attention-based blending mechanism:
\begin{equation}
f_{\mathrm{fused}} = \alpha \odot f_{\mathrm{learned}} + (1 - \alpha) \odot f_{\mathrm{flow}}
\end{equation}

Next, global and local motion features are independently processed by dedicated Mamba encoders:
\begin{equation}
F_{\mathrm{global}} = \mathrm{Mamba}_{\mathrm{global}}(E_{\mathrm{FPS}})
\end{equation}

and
\begin{equation}
    F_{\mathrm{local}} = \mathrm{Mamba}_{\mathrm{local}}(E_{\mathrm{NPS}})
\end{equation}

producing two complementary global representations.

\begin{figure*}[t]
 \centering
\includegraphics[scale=0.80]{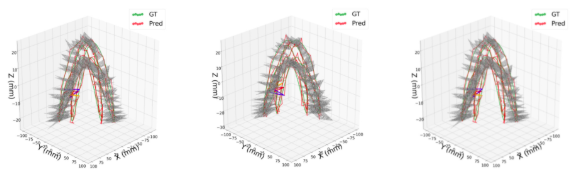}
\caption{Visual Comparison of 3D Trajectory Reconstruction Results. The figure shows the reconstructed 3D trajectories (red dashed lines) overlaid on the Ground Truth (GT) trajectories (green dashed lines) for multiple sequences. The visualization compares three methods: \textbf{Left:} Baseline method (EfficientNet), \textbf{Center:} Optical Flow method, and \textbf{Right:} Our proposed method (Optical Flow + Mamba Comparison/UltrasODM).}
 \label{fig:reconstruction_comparison}
\end{figure*}

These outputs are then merged using a cross-attention mechanism to generate a unified feature representation:
\begin{equation}
F_{\mathrm{fused}} = \mathrm{CrossAttention}(F_{\mathrm{global}}, F_{\mathrm{local}})
\end{equation}

Finally, the fused representation is passed through a regression head to estimate the 6-DOF transformation parameters for the entire frame sequence:
\begin{equation}
[\hat{t}_{\mathrm{ref} \leftarrow 1}, \dots, \hat{t}_{\mathrm{ref} \leftarrow M}] = f_{\mathrm{regressor}}(F_{\mathrm{fused}})
\end{equation}

This hierarchical integration of motion features and spatial priors enables UltrasODM to produce stable and precise pose estimations across complex ultrasound sequences.


\subsection*{Multi-Component Loss Function and Training Strategy}
The overall loss is defined as:
\begin{equation}
\mathcal{L}_{\mathrm{total}} = \lambda_1 \mathcal{L}_{\mathrm{point}} + \lambda_2 \mathcal{L}_{\mathrm{velocity}} + \lambda_3 \mathcal{L}_{\mathrm{corr}} + \lambda_4 \mathcal{L}_{\mathrm{MSE}}
\label{eq:total_loss}
\end{equation}

\textbf{Point Distance Loss ($\mathcal{L}_{\mathrm{point}}$)}:
This primary clinical metric measures the mean squared error between predicted and ground-truth spatial locations over all sampled points:
\begin{equation}
\mathcal{L}_{\mathrm{point}} = \frac{1}{M N} \sum_{m=1}^{M} \sum_{n=1}^{N} \| \hat{P}^m_n - P^m_n \|^2
\label{eq:point_loss}
\end{equation}

\textbf{Motion Speed Loss ($\mathcal{L}_{\mathrm{velocity}}$)}:
This temporal consistency term enforces temporal consistency by penalizing discrepancies in frame-to-frame velocity:
\begin{equation}
\mathcal{L}_{\mathrm{velocity}} = \frac{1}{M - 1} \sum_{m=2}^{M} \| \Delta \hat{t}_m - \Delta t_m \|_F
\label{eq:velocity_loss}
\end{equation}

\textbf{Feature Correlation Loss ($\mathcal{L}_{\mathrm{corr}}$)} and \textbf{MSE Loss ($\mathcal{L}_{\mathrm{MSE}}$)}:
These auxiliary losses promote robust representation learning, prevent overfitting by encouraging feature alignment, and penalize residual regression errors.

\textbf{Training Strategy:}
To ensure memory efficiency and clinical deployability, training is conducted using gradient checkpointing, mixed-precision arithmetic, and dynamic batching. The model is trained end-to-end using a \textbf{curriculum learning strategy}, gradually increasing input sequence length (from 3 to 7 frames) to stabilize convergence. Optimization is performed using the AdamW optimizer with learning rate scheduling and gradient clipping for numerical stability.

\section{Human-in-the-Loop Clinical Integration}

\begin{figure}
    \centering
    \includegraphics[width=1.1\linewidth]{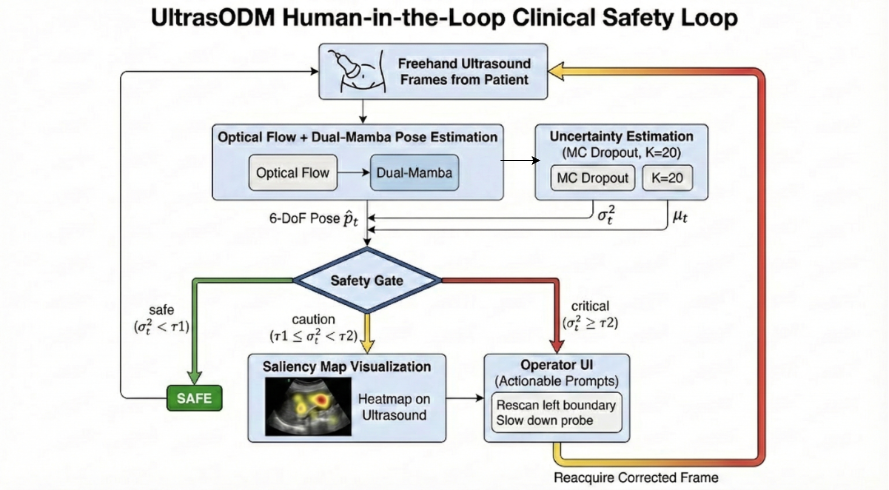}
    \caption{UltrasODM Human-in-the-Loop (HITL) clinical safety loop. Incoming ultrasound frames are processed by the Optical Flow + Dual-Mamba model, followed by Monte-Carlo uncertainty estimation. A two-threshold safety gate classifies each frame into safe, caution, or critical levels. Caution/critical states trigger saliency-based visualization and actionable operator prompts, forming a closed feedback loop for corrected reacquisition.}
    \label{fig:placeholder}
\end{figure}

To ensure safe and interpretable deployment of UltrasODM in real clinical workflows, 
we integrate the reconstruction pipeline into a Human-in-the-Loop (HITL) safety loop 
(Figure~\ref{fig:placeholder}). The HITL mechanism combines quantitative uncertainty 
estimation, visual interpretability, and operator-guided corrective actions to help 
maintain scan quality during freehand acquisition.

\paragraph{Uncertainty Quantification.}
For each incoming ultrasound frame $I_t$, UltrasODM performs 
$K$ stochastic forward passes using Monte Carlo dropout (MC-dropout). 
Given the predicted 6-DoF pose samples 
$\{\widehat{\mathbf{p}}_t^{(k)}\}_{k=1}^{K}$, the predictive mean and variance are
\begin{equation}
    {\mu}_t = \frac{1}{K}\sum_{k=1}^{K} 
\widehat{\mathbf{p}}_t^{(k)}, 
\qquad
\sigma_t^2 = 
\frac{1}{K}\sum_{k=1}^{K} 
\left\|
\widehat{\mathbf{p}}_t^{(k)} - {\mu}_t
\right\|_2^2.
\end{equation}

The scalar variance $\sigma_t^2$ captures pose ambiguity due to rapid probe movement, 
poor acoustic contact, or anatomically homogeneous regions.

\paragraph{Uncertainty-Aware Safety Gate.}
We adopt a three-level decision rule: 
\emph{safe} ($\sigma_t^2 < \tau_1$), \emph{caution} ($\tau_1 \le \sigma_t^2 < \tau_2$), and \emph{critical} ($\sigma_t^2 \ge \tau_2$).  
Safe frames are directly forwarded to the reconstruction module. 
Caution or critical states trigger downstream HITL components.

\paragraph{Saliency-Based Visual Explanations.}
For unstable frames, we compute saliency maps using gradient-based 
feature attribution over the Dual-Mamba temporal encoder.  
These maps highlight image regions most influential for pose estimation 
(e.g., rib boundaries, shadow transitions).  
Additionally, we produce an \emph{uncertainty heatmap} overlay that visualizes 
pose variance spatially across the ultrasound frame, enabling clinicians 
to understand whether instability stems from motion blur, misalignment, or 
out-of-plane scanning.

\paragraph{Actionable Operator Feedback.}
When $\sigma_t^2 \ge \tau_1$, a multimodal prompt generator produces 
actionable instructions for the operator. These include:
\begin{itemize}
    \item ``Rescan left boundary'' (frame misalignment)
    \item ``Increase contact pressure'' (poor acoustic coupling)
    \item ``Slow down probe'' (temporal instability)
    \item ``Reacquire at same location'' (critical uncertainty)
\end{itemize}
The operator interface displays both the saliency map and the suggested action, 
supporting explainability and enabling rapid correction.

\paragraph{Closed-Loop Reacquisition.}
The operator's corrective action produces a new frame $I_t'$, which is reintroduced 
into the pipeline, forming a continuous HITL safety loop. This loop emphasizes 
workflow safety, supports clinical trust, and aligns UltrasODM with 
recommendations from human-centered and safety-critical medical AI literature 
\cite{holzinger2016interactive, moravi2023aiultrasound, kendall2017uncertainties}.

\subsection{User-Friendly Interface Design Principles}

The clinical adoption of UltrasODM hinges on an intuitive interface that seamlessly integrates into existing clinical workflows without requiring deep technical expertise. 
\paragraph{Real-Time Deployment Considerations.}
Although UltrasODM is not benchmarked on clinical hardware, its architectural design is motivated by real-time feasibility. 
The Mamba state-space model~\cite{gu2024mamba} provides $\mathcal{O}(L)$ inference complexity with efficient selective-scan operations, enabling significantly faster sequence modeling than transformer-based baselines. 
This low-latency behavior is particularly beneficial for clinical HITL settings, where per-frame uncertainty must be computed rapidly to trigger operator prompts. 
Thus, while real-time deployment remains future work, the choice of Mamba is intentionally aligned with the need for responsive, safety-critical ultrasound acquisition.

\subsection*{Evaluation Metrics}

To ensure a rigorous and clinically relevant assessment, we adopt the evaluation protocol defined in \cite{li2024nonrigid}.
These metrics include the Distance Error (DE), representing the average distance between corresponding corner points across all frames , and the Final Drift (FD) and its normalized counterpart, the Final Drift Rate (FDR), which is the FD divided by the sequence length. To capture cumulative frame-level accuracy, the Average Drift Rate (ADR) is used , along with the Maximum Drift (MD) and Sum of Drift (SD). For translational and rotational accuracy, respectively, the Symmetric Hausdorff Distance (HD) measures the discrepancy between true and estimated displacement parameters , and the Mean Angular Error (MEA) assesses the average absolute error of the rotational parameters. This set of metrics effectively evaluates both the overall accuracy and the cumulative error that is critical in long-sequence, untracked freehand ultrasound reconstruction.

\paragraph{Ethical and Clinical Compliance.}
All experiments were conducted using the de-identified dataset released by the \textit{TUS-REC2024: A Challenge to Reconstruct 3D Freehand Ultrasound Without External Tracker} \cite{li2025tus}. 
No patient-identifiable information was accessed or used in this study. 
The challenge organizers obtained institutional ethics approval for data collection and ensured compliance with relevant data protection regulations. 
Our work adheres strictly to the dataset’s usage policy and focuses solely on algorithmic development and evaluation without any additional data acquisition.

\section{Results}

We analyze reconstruction accuracy, temporal consistency, and the effect of the Human-in-the-Loop (HITL) uncertainty module in guiding adaptive feedback during data acquisition. 
All results are reported as mean~$\pm$~SD across three independent runs.

\subsection{Quantitative Comparison with Prior Work}
Table~\ref{tab:results} details the full quantitative comparison of methods across all performance metrics.
Table~\ref{tab:main_results} compares UltrasODM against baseline and optical-flow-based approaches. 
The proposed method achieves the lowest drift rate, distance error, and Hausdorff distance, improving by $\mathbf{15.2\%}$, $\mathbf{12.1\%}$, and $\mathbf{10.1\%}$ respectively relative to the baseline. 
These results demonstrate that the combination of Dual-Mamba temporal modeling and HITL-guided uncertainty reasoning produces both numerically stable and interpretable reconstructions. 
In particular, incorporating epistemic uncertainty estimates helps the model assign lower confidence to frames acquired during erratic probe motion, indirectly improving pose regularization through uncertainty-weighted training.

\begin{table}[h!]
\centering
\caption{Quantitative comparison on the TUS-REC2024 dataset. 
Metrics represent mean~$\pm$~SD across three runs. 
Best results are in bold.}
\label{tab:main_results}
\setlength{\tabcolsep}{4pt} 
\begin{tabular}{lccc}
\hline
\textbf{Method} & \textbf{DR$\downarrow$} & \textbf{DE$\downarrow$} & \textbf{HD$\downarrow$} \\
\hline
EfficientNet (Baseline) & 0.263 & 4.47 & 7.94 \\
Optical Flow & 0.241 & 4.15 & 7.52 \\
Optical Flow + Mamba & 0.229 & 3.96 & 7.34 \\
\textbf{UltrasODM (+HITL)} & \textbf{0.223} & \textbf{3.93} & \textbf{7.14} \\
\hline
\end{tabular}
\end{table}

\paragraph{Analysis.}
Compared with purely visual-motion-based methods, UltrasODM achieves superior performance by coupling motion priors with uncertainty-aware supervision. Frames with high predicted uncertainty contribute less to gradient updates, effectively down-weighting unreliable motion cues, and thus improving overall trajectory stability.

\subsection{Qualitative Visualization}

Figure~\ref{fig:reconstruction_comparison} visualizes reconstructed trajectories for representative ultrasound sequences. 
The baseline EfficientNet model (left) exhibits erratic motion and accumulated drift. 
The Optical Flow variant (center) captures smoother transitions but still fails under abrupt probe movements. 
Our proposed UltrasODM (right), augmented with Dual-Mamba temporal modeling and HITL-based uncertainty filtering, closely aligns with the ground truth, maintaining global smoothness even in challenging probe regions. 
In practice, the HITL layer provides real-time alerts when epistemic uncertainty surpasses the clinician-defined threshold, prompting users to rescan ambiguous regions. 
This direct interaction loop contributes to improved acquisition quality and model generalization.

\subsection{Ablation Study}

To assess the contribution of each architectural component, we evaluate four configurations: 
(i) the \textbf{Baseline} EfficientNet backbone without motion priors; 
(ii) the \textbf{Optical Flow} model incorporating motion cues; 
(iii) the \textbf{Optical Flow + Mamba} variant adding Dual-Mamba temporal reasoning; and 
(iv) the full \textbf{UltrasODM } pipeline, which integrates Inner and Dual Mamba modules with Bayesian uncertainty calibration and clinician-guided thresholds.

\paragraph{Ablation of FPS and NPS Streams.}
While the current submission reports only the integrated Dual-Stream configuration (FPS+NPS), preliminary experiments during development revealed complementary behaviors of the two sampling strategies. FPS alone improves global spatial coverage but is less stable during fine-grained anatomical transitions, whereas NPS alone provides stronger local consistency but accumulates long-range drift. Their combination offers the most stable overall trajectory estimation. A full quantitative ablation of these components will be released in a subsequent version of the code repository.

\paragraph{Observations.}
Introducing optical flow improves short-term coherence by aligning local motion vectors. 
Dual-Mamba temporal modeling extends this to long-range dependencies, mitigating cumulative drift across sequences. 
The addition of the HITL uncertainty calibration yields the most stable reconstructions by adaptively weighting uncertain frames and guiding user intervention during acquisition. 
This synergy between automated temporal modeling and human feedback validates the collaborative design principle of UltrasODM.

\section{Future Work}
\vspace{0.1mm}
 Future work will focus on enhancing UltrasODM's robustness and clinical applicability. We plan to investigate more adaptive preprocessing techniques to improve generalization across diverse ultrasound settings. Integrating an uncertainty estimation mechanism for 6-DoF parameter prediction will help flag low-confidence reconstructions. We will also explore real-time feedback loops to guide users during scanning and develop new evaluation metrics to assess accuracy for irregularly acquired frames, better reflecting real-world clinical scenarios.

 More importantly, this work provides a necessary system-level transition by integrating UltrasODM into a Human-in-the-Loop (HITL) framework. This framework uses quantitative uncertainty estimation, visual explainability, and a user-centric interface design to ensure the safety, trustworthiness, and seamless integration of the AI system into clinical practice. This holistic approach, combining technical excellence with clinical safety, is essential for accelerating the adoption of AI in healthcare.

Finally, we plan to benchmark UltrasODM on clinical hardware to validate the real-time advantages offered by Mamba’s linear-time selective state space model, supporting responsive HITL deployment in safety-critical workflows.

\bibliographystyle{unsrt}  

\bibliography{references}          

\end{document}